\documentclass[journal]{IEEEtran}

\ifCLASSINFOpdf
  \usepackage[pdftex]{graphicx}
\else
  \usepackage[dvips]{graphicx}
\fi
\usepackage{amsmath}
\usepackage{amsfonts}
\usepackage{amssymb}
\usepackage{amsthm}
\usepackage{algorithmic}
\usepackage{algorithm}
\usepackage[export]{adjustbox}
\usepackage{booktabs}

\usepackage{array}
\usepackage{tabularx}
\usepackage{multirow}
\usepackage{rotating}
\usepackage{array}

\usepackage[caption=false,font=footnotesize]{subfig}

\usepackage{changepage}

\usepackage{url}
\usepackage{siunitx}
\begin{document}

\title{A 2D Sinogram-Based Approach to Defect Localization in Computed Tomography}

\author{Yuzhong~Zhou,
        Linda-Sophie~Schneider,
        Fuxin Fan,
        Andreas Maier\thanks{All authors are with the Pattern Recognition Lab, Friedrich-Alexander-Universit\"at Erlangen-N\"urnberg, Erlangen, Germany}}

\maketitle

\begin{abstract}
  The rise of deep learning has introduced a transformative era in the field of image processing, particularly in the context of computed tomography. Deep learning has made a significant contribution to the field of industrial Computed Tomography. However, many defect detection algorithms are applied directly to the reconstructed domain, often disregarding the raw sensor data. This paper shifts the focus to the use of sinograms. Within this framework, we present a comprehensive three-step deep learning algorithm, designed to identify and analyze defects within objects without resorting to image reconstruction. These three steps are defect segmentation, mask isolation, and defect analysis. We use a U-Net-based architecture for defect segmentation. Our method achieves the Intersection over Union of 92.02\% on our simulated data, with an average position error of 1.3 pixels for defect detection on a 512-pixel-wide detector.
\end{abstract}

\begin{IEEEkeywords}
Computed Tomography, image segmentation, defect localization, sinogram
\end{IEEEkeywords}

\IEEEpeerreviewmaketitle

\section{Introduction}

\IEEEPARstart{C}{omputed} Tomography (CT) is a remarkable technology that produces highly detailed cross-sectional images that provide valuable insights into the internal structures of objects, such as industrial components or the human body in the medical field. These images are essential for detecting abnormalities, which can range from medical conditions such as tumors and fractures in the human body to imperfections or defects in industrial products.

Detecting defects in CT images is of key importance in a variety of applications. In the medical field, early detection of medical conditions can be a determining factor for successful treatment and patient outcomes. Deep learning, a subset of machine learning, has emerged as a transformative force in CT image analysis. Much research has shown that convolutional neural networks can significantly improve the performance of defect detection~\cite{usman2020volumetric, dong2020multi}, especially semantic segmentation networks such as fully convolutional neural networks \nocite{long2015fully} and U-Net\nocite{ronneberger2015u}.

Most research about defect detection focuses on image processing operations performed on reconstructed domains. Therefore, regardless of how efficient these algorithms are, they must wait for the completion of the reconstruction before they can start. In particular, prominent reconstruction algorithms such as iterative reconstruction (IR) \nocite{herman2009fundamentals} introduce significant computational overhead. As shown in previous work \cite{pham2023accurate}, the IR algorithm, called real space iterative reconstruction, requires approximately 30 seconds to reconstruct a 3D volume measuring $243\times243\times243$. Consequently, the time required for defect detection in this pipeline is at least 30 seconds.

These studies based on the reconstructed domain overlook the valuable information in the raw sensor data, represented as sinograms---X-ray attenuation measurements taken from different angles. Traditional image reconstruction from sinograms, while standard, has drawbacks such as loss of detail, the introduction of artifacts, and high computational demands\nocite{wang2016image}. 

This research focuses on defects that have high attenuation. We exploits the potential of sinograms in defect detection by proposing direct analysis in sinogram space. This could allow simultaneous defect detection and reconstruction, or even skip reconstruction when unnecessary.
This work aims to develop a methodology for defect detection and localization in parallel beam computed tomography using a U-Net-based architecture tailored for the analysis of CT sinograms. Our approach, which is divided into sinogram segmentation, instance segmentation, and defect analysis, can detect defects directly within sinograms, bypassing the need for reconstruction. This method detects defects efficiently and can be integrated with reconstruction processes in parallel beam CT. With this approach, we have not only improved defect detection in computed tomography, but also significantly reduced the computational resources required, paving the way for more efficient and accurate defect detection.

\section{Methods}
\subsection{Sinogram Segmentation}
\begin{figure}[!t]
  \centering
  \includegraphics[width=2.5in]{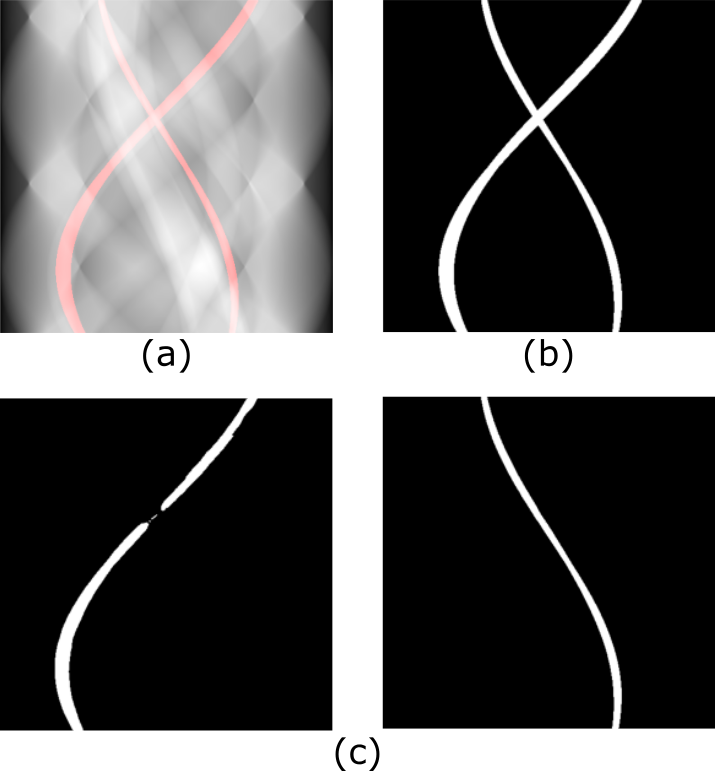}
  \caption{(a) Input of the U-Net, pixels containing information of defects are painted red; (b) Result of sinogram segmentation is a binary mask that contains information about all found defects; (c) Result of instance segmentation, the information in (b) is separated based on which defect it belongs to.}
  \label{fig: seg}
\end{figure}

Sinogram segmentation, which is a semantic segmentation task, is initially addressed using a U-Net model. This model takes the sinogram of the object (Fig. \ref{fig: seg}(a)) and processes it through 3 encoder layers for downsampling and feature extraction, and 3 decoder layers for upsampling and classification of pixels. Starting with 32 feature channels, each layer of the U-Net doubles the channels of the previous layer. This U-Net performs a binary classification of the sinogram pixels, identifying abnormal pixels as foreground and others as background. The goal is to isolate defect information in the sinogram mask (Fig. \ref{fig: seg}(b)). Our architecture is rigorously trained, with continuous validation using a separate dataset to monitor metrics such as Intersection over Union (IoU), precision, recall, and F1-score.

\subsection{Instance Segmentation}
Sinogram segmentation produces a binary mask that contains all defects. To analyze these defects individually, we implement instance segmentation, which, unlike semantic segmentation, distinguishes separate defects within the same mask (Fig. \ref{fig: seg}(c)). In Fig. \ref{fig: seg}(b), the distinct positioning of two defects in the volume leads to separate projections in the sinogram. Using the Radon transform, identifying the central sinusoid of each defect allows them to be separated. This is achieved through skeletonization and reclassification techniques.

\subsubsection{Skeletonization}
\begin{figure}[!t]
  \centering
  \includegraphics[width=3.5in]{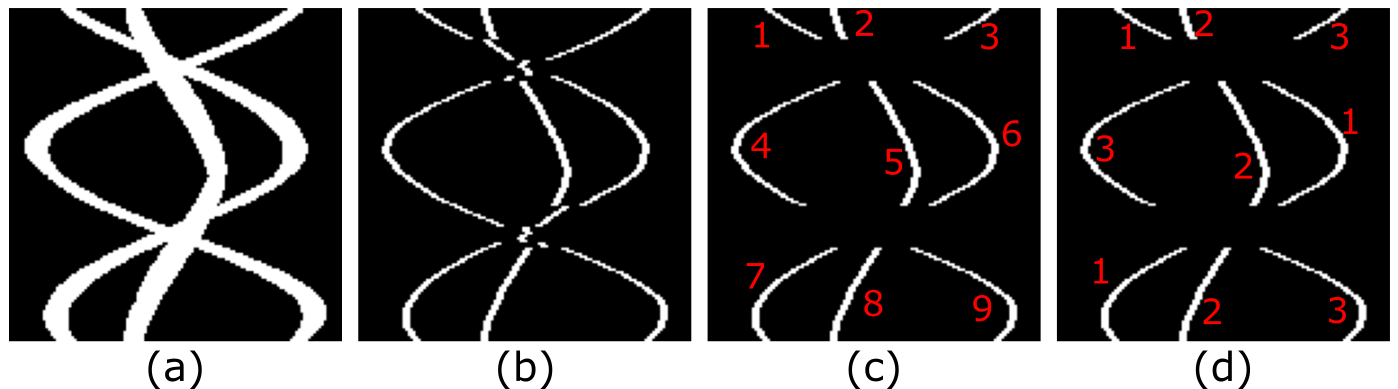}
  \caption{(a) Result of semantic segmentation, which is also the input of the instance segmentation algorithm; (b) Raw result of skeletonization. After erasing the intersection part we get the final skeletonization result (c). (d) is the result of reclassification. For ease of presentation, the results of the skeletonization in (b), (c), and (d) have been widened.}
  \label{fig: ske}
\end{figure}

we address the clustering of defect projections by adapting the Zhang-Suen algorithm \cite{zhang1984fast} to the characteristics of the Radon transform.  In each row of the sinogram, we reduce every non-zero region to its central point, producing an initial skeletonization result (Fig. \ref{fig: ske}(b)). Next, intersections are removed, resulting in distinct paths (Fig. \ref{fig: ske}(c)), each path containing information about a specific defect.
 
\subsubsection{Reclassification}
In Fig. \ref{fig: seg}(c), the original mask is thinned out, with the remaining pixels representing the centers of each defect's projection in a given direction. These pixels are crucial for computing the sinusoids needed to differentiate the defects. Given our projection's known frequencies and offsets, we only need to determine amplitude and phase. We choose not to use the Hough transform directly for parameter extraction; instead, we first categorize these pixels based on Radon transform characteristics. Pixels in the same group contribute to the calculation of an amplitude-phase pair, which is then fed into the Hough transform to derive the necessary parameters. This process is detailed in Algorithm \ref{connection}.
\begin{algorithm}
  \caption{Reclassification}
  \label{connection}
  \begin{algorithmic}
    \REQUIRE Result of skeletonization $M_s$
    \ENSURE Parameters for separation
    \STATE Initialize $M_l$ as a zero matrix
    \STATE Initialize working row $r$ to zero
    \STATE Initialize start of label $s$ to 1
    \STATE Initialize end of label $e$ to 1

    \WHILE{$r$ is less than the height of $M_s$}
      \STATE $idx$ = indices of nonzero elements in $M_s[r]$
      \IF{$idx$ is empty}
        \STATE $s$ = $e$
      \ELSE
        \STATE $e$ = $s$ + size of $idx$
        \STATE $M_l[r, idx]$ = range from $s$ to $e$
      \ENDIF
      \ENDWHILE

      \STATE Apply Hough Transform for sinusoid on labeled paths
      \STATE Re-label based on the result of Hough Transform
  \end{algorithmic}
\end{algorithm}

In the final step, the skeletons isolated with different labels as shown in Fig. \ref{fig: ske}(d) are merged with the sinogram mask to create distinct sinogram masks for each defect. We use the stored radius of non-zero regions to dilate the skeleton accordingly.

\subsection{Defect Analysis}
When X-rays are projected onto a detector along a single beam direction, the resulting projection width can be used to define an area encompassing the object. In two dimensions, performing a similar operation in a different direction produces a different area, the intersection of which indicates the object's position. This intersection forms the minimum enclosing parallelogram of the object, and the center is found at the intersection of the midlines of these two projections. These midlines are what we get from the reclassification, which gives the coordinates of the center. The lengths of these projections allow the overlapping area to be calculated and the size of the defect to be estimated.

\begin{figure}[!t]
  \centering
  \includegraphics[width=3in]{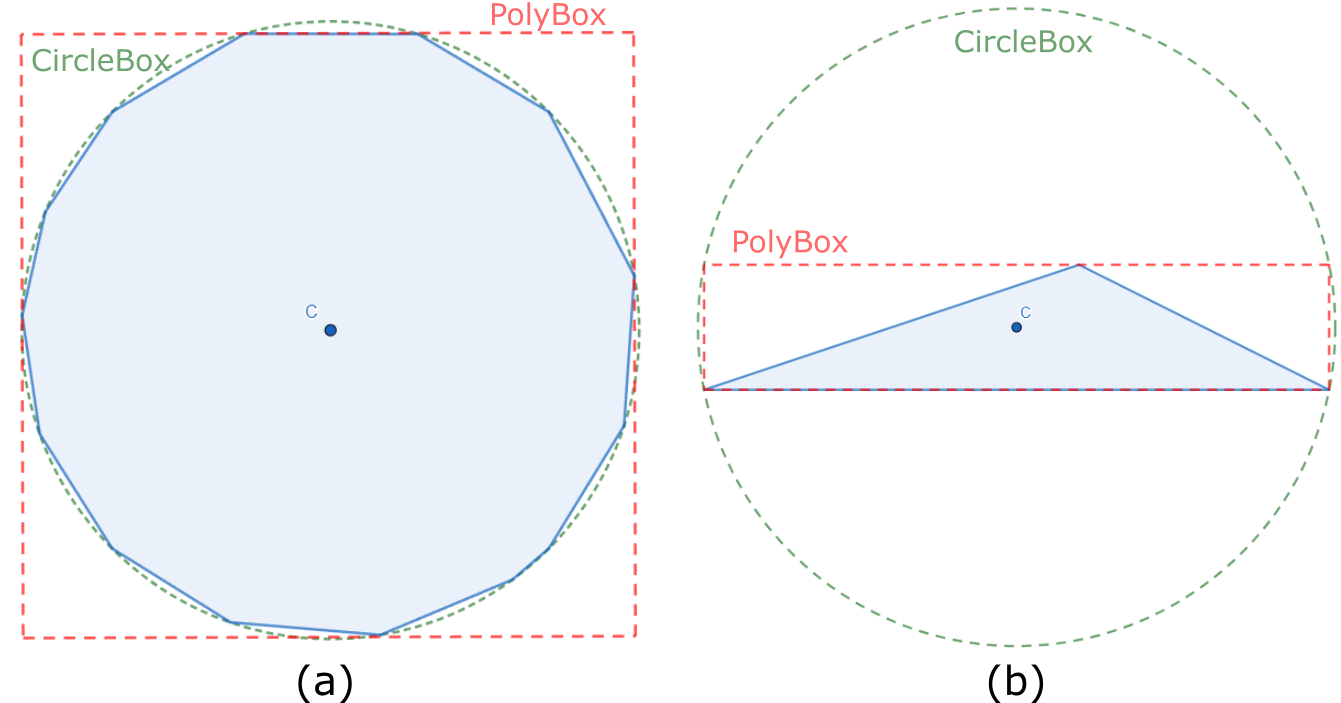}
 \caption{Appropriate strategies vary with defect shape. (a) For circular defects, the CircleBox method outperforms the overlap method of two different projections in terms of accuracy. (b) For non-circular defects, the CircleBox approach is inappropriate.}
  \label{fig: ana}
\end{figure}

However, it's important to note that with different pairs of projections, we can have multiple enclosing parallelograms, possibly with different centers. For non-circular defects, one approach is to pair projections and average the results, but this increases computational complexity. Alternatively, the computation can focus on the shortest projection and its perpendicular counterpart. For circular defects, the centerlines typically intersect at the center of the circle. Here, the properties of the Radon transform indicate that the center aligns with a sine wave created by the projections of the center points of the sinogram. Once the sinogram mask for the defect is obtained, this sine wave can be calculated to determine the center coordinates of the circle, as shown in Figure \ref{fig: ana}.

\section{Experiment}
The U-Net architecture was developed using Pytorch\nocite{paszke2019pytorch} and Pytorch Lightning\nocite{falcon2019pytorch}. Network parameters were randomly initialized, followed by training using backpropagation and the Adam optimizer\nocite{kingma2014adam}, starting with a learning rate of 0.001. The system was operated on Windows 10, using an NVIDIA 3060 graphics card.
\begin{figure}[!t]
  \centering
  \includegraphics[width=3.4in]{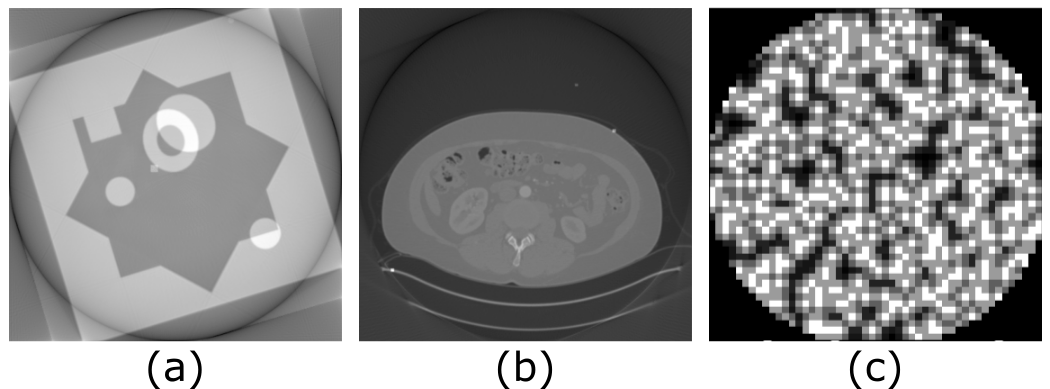}
  \caption{(a) Sample from the custom dataset. (b) Sample from KiTS23 kidney dataset. (c) Sample of a defect used in this paper which has a radius of 10 pixels.}
  \label{fig: data}
\end{figure}

We used two different datasets with two different types of volumes as the background: a custom MagicCube and the KiTS23 volumes \cite{heller2023kits21} (see Fig. \ref{fig: data}). The custom MagicCube contains various shapes and forms. The KiTS23 volumes are reconstructed patient volumes. For this study's focus on 2D parallel ray projections, we applied this technique to each slice of both volume types, creating simulated 2D sinogram datasets.

For the custom MagicCube, simulated sinograms were generated using the PYRO-NN toolkit \cite{syben2019pyro} for 2D parallel ray projection. The sinograms covered a rotation range from 0 to 180 degrees, with a total of 512 sinograms and a 1D detector size of 512 pixels. The dimensions of the MagicCube are $512\times512\times512$, which corresponds to 512 slices of 2D phantoms, each $512\times512$ in size. We allocated 80\% of these slices for training, and the remainder for validation and testing. 

For data augmentation, one defect was introduced in each 2D phantom, and the training phantoms were rotated between 0 and 90 degrees in 1-degree increments. This process resulted in 40,500 sinograms for training, each containing a single defect, and 5,580 for testing, each containing up to three defects.

For defect simulation, we manually introduced two types of defects: small patches of circular or square Gaussian noise, as shown in \ref{fig: data}(c). The radius of these patches is randomly chosen from 8 to 30 pixels. During training data generation, the original images were randomly rotated, and the defects were randomly positioned within these images to ensure no overlap in the case of multiple targets. Each training set used only one defect type. Because our defects were artificially added, generating label masks was straightforward. We simply calculated the difference between the original object and the object with the added defects to generate the label masks.

We evaluated our pipeline using different metrics tailored to each step. For sinogram segmentation, the impact of false positives on subsequent instance segmentation algorithms was significant. To address this, IoU and precision were chosen as the primary metrics. 

Additionally, recall and F1 score were considered as supplementary metrics. When IoU values were comparable, recall was useful for identifying more true positives, and F1-score assisted in balancing precision and recall at the pixel level.

\section{Results}
\begin{figure}[!t]
  \centering
  \includegraphics[width=3.4in]{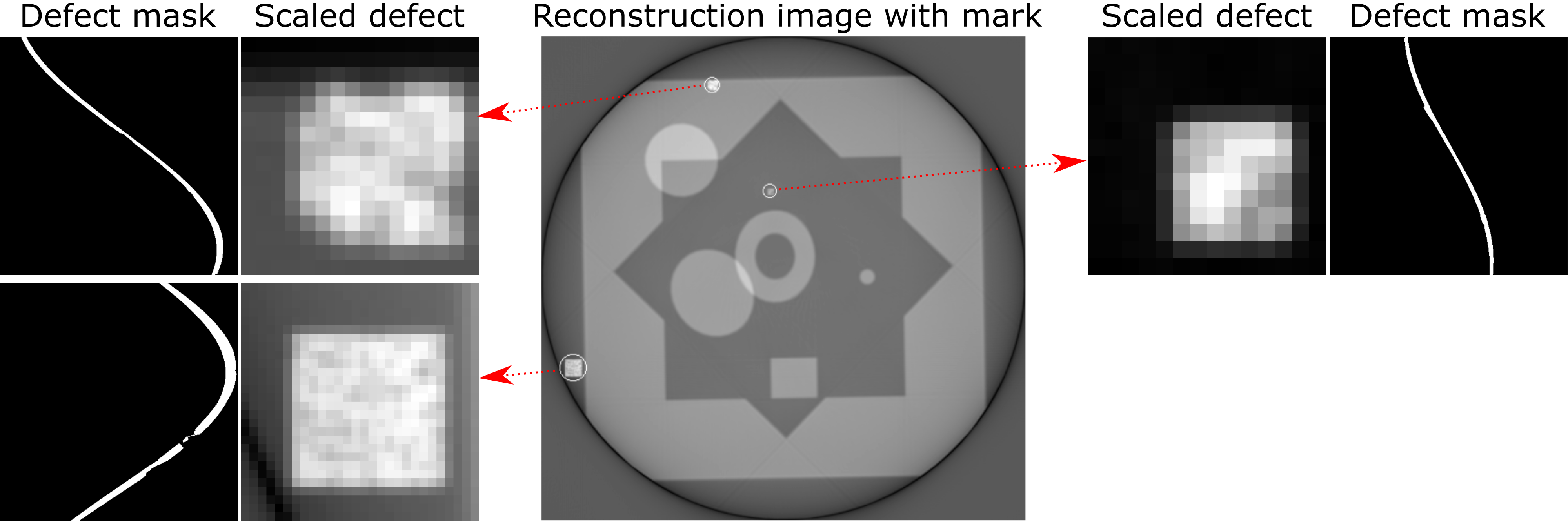}
  \caption{Our method can detect and analyze the defects before reconstruction and facilitate defect identification in reconstructed domain.}
  \label{fig: res}
\end{figure}
We first demonstrate the practical effectiveness of our method for extracting defect information from sinograms, as shown in Fig. \ref{fig: res}. This figure effectively demonstrates the successful isolation of three different defects within a sinogram. Furthermore, by analyzing the sinogram mask individually, we can determine the positions and sizes of these defects, which will aid in their identification in the reconstructed domain, and importantly, this analysis is performed prior to image reconstruction. The results in Fig. \ref{fig: res} confirm the achievement of our goal.

Since our method involves three steps, each of which not only affects the others, but also provides opportunities for isolated evaluation, we have performed tests on the entire process as well as on each individual step. The results of these tests are detailed in Table \ref{tab: seg} and Table \ref{tab: ana}.  
\begin{table}[ht]
  \caption{Sinogram Segmentation Metrics}
  \label{tab: seg}
  \centering
  \begin{tabular}{ccccc}
    \hline
    \bfseries Dataset & \bfseries IoU & \bfseries Precision & \bfseries Recall & \bfseries F1 \\
    \hline
    MagicCube & 0.9202 & 0.9992 & 0.9262 & 0.9572 \\
    KiTS23 & 0.8172 & 0.9943 & 0.8200 & 0.8888 \\
    \hline
  \end{tabular}
\end{table}
\ifCLASSOPTIONcaptionsoff
  \newpage
\fi

As shown in the results, our method shows a remarkably high average IoU of 0.9202 on the MagicCube-derived sinograms. This high IoU value indicates the efficiency of our method in accurately isolating the defect-containing portions from the sinogram data.
\begin{table}[ht]
  \caption{Instance Segmentation Metrics}
  \label{tab: ana}
  \centering
  \begin{tabular}{p{0.5in}p{0.7in}p{0.7in}p{0.7in}}
    \hline
    \bfseries Dataset & \bfseries Correct rate & \bfseries Distance \newline relative error & \bfseries Area  \newline relative error \\
    \hline
    MagicCube & 92.5\% & 0.0055 & 0.0197 \\
    KiTS23 & 70.7\% & 0.0054 & 0.0139 \\
    \hline
  \end{tabular}
\end{table}
\ifCLASSOPTIONcaptionsoff
  \newpage
\fi

In the following instance segmentation step, we introduced a metric named segmentation rate, which means the percentage of correct instance segmentation. We achieve an accurate segmentation rate of 92.5\%. Errors are mainly due to defect overlap or inaccuracies in the results of the previous step. For defect localization, we use the relative error to show the accuracy, the relative error is the ratio of the pixel error to the width of the detector. The average relative error is 0.0055, implying that on a 512-pixel wide detector, our method computes an average error of about 1.3 pixels for defect positions. The average relative error for the radius is 0.0197.

On the KiTS23 dataset, the performance is slightly worse, with an average IoU of only 0.8172 and an accurate segmentation rate of 70.7\%. This is mainly due to the more complex pixel distribution in medical images. The most severe impact comes from the first step of semantic segmentation, as subsequent steps rely on the previous step's results. Consequently, there is a slight decrease in the overall performance.

\section{Discussion and Conclusion}
This work presents an innovative approach to defect detection and localization in parallel beam computed tomography using a specialized U-Net-based architecture tailored for the analysis of CT sinograms. This methodology consists of sinogram segmentation, instance segmentation, and defect analysis, allowing for direct detection of defects within sinograms, and eliminating the need for image reconstruction.

Our method not only efficiently detects defects, but also offers the potential for integration with reconstruction processes in parallel beam CT. As a result, we have shown the average IoU of our method can be 0.9202, with an average position error of 1.3 pixels for defect detection on a 512-pixel-wide detector. 

Experimental results demonstrate the effectiveness of our approach, with high IoU values in sinogram segmentation and accurate defect localization even in complex scenarios. While the performance on the KiTS23 dataset showed a slight decrease due to the challenging nature of medical images, the overall impact of our method on the efficiency and accuracy of defect detection is still remarkable.

Currently, our method is tailored for 2D parallel ray projection analysis. Due to the limitations of our current experimental setup, we have used only simulated projection data and have omitted physical projection testing. Our primary focus for future research is to extend our methodology from 2D to 3D analysis and to transition from parallel to cone beam projection scenarios.

\bibliographystyle{IEEEtran}

\bibliography{refer}

\end{document}